\def\year{2019}\relax
\documentclass[letterpaper]{article} 
\usepackage{aaai19}  
\usepackage{times}  
\usepackage{helvet}  
\usepackage{courier}  
\usepackage{url}  
\usepackage{graphicx}  
\usepackage{algorithmicx} 
\usepackage{amsmath} 
\usepackage{mathtools} 
\usepackage{amsfonts} 
\usepackage{verbatim} 
\usepackage{tikz}

\DeclarePairedDelimiter{\ceil}{\lceil}{\rceil} 
\newcommand{\commentedbox}[2]{%
  \mbox{
    \begin{tabular}[t]{@{}c@{}}
    $\boxed{\displaystyle#1}$\\
    #2
    \end{tabular}%
  }%
}

\begin{document}
%
\title{Fuzzy Hashing as Perturbation-Consistent Adversarial Kernel Embedding}
\author{Ari Azarafrooz \and John Brock \\
Cylance, Department of Research and Intelligence.\\
Irvine, CA, USA}
\maketitle
\begin{abstract}
Measuring the similarity of two files is an important task in malware analysis, with fuzzy hash functions being a popular approach. Traditional fuzzy hash functions are data agnostic: they do not learn from a particular dataset how to determine similarity; their behavior is fixed across all datasets. In this paper, we demonstrate that fuzzy hash functions can be learned in a novel minimax training framework, and that these learned fuzzy hash functions outperform traditional fuzzy hash functions at the file similarity task for Portable Executable files. In our approach, hash digests can be extracted from the kernel embeddings of two kernel networks, trained in a minimax framework, where the roles of players during training (i.e adversary versus generator) alternate along with the input data. We refer to this new minimax architecture as \textit{perturbation-consistent}. The similarity score for a pair of files is the utility of the minimax game in equilibrium. Our experiments show that learned fuzzy hash functions generalize well, capable of determining that two files are similar even when one of those files was generated using \textit{insertion} and \textit{deletion} operations.
\end{abstract}

\section{Introduction}

\subsection{File Similarity in Malware Analysis}
The rapid proliferation of malware poses a substantial set of technical and organizational challenges for malware analysts. For example, for the year 2017, there were more than 121.6 million instances of new malware recorded, i.e., on average there were about 3.9 new malware instances being created every second (AV-TEST 2018). While these numbers may seem large, it is important to note that most new malware is not written from scratch. Instead, most new malware files are only slightly modified versions of previously known malware. New malware files will often be near-exact duplicates of previously seen malware. There are several reasons for this near-duplication, including malware authors making small changes to bypass new malware defenses, as well as code reuse and code sharing among malware authors (Walenstein and Lakhotia 2007). 

Given some unknown file, two important questions a malware analyst might ask are: Is this new file benign or malicious; and, if the file is malicious, what kind of malware is it (e.g., ransomware, Trojan, credential stealer, etc.)? Malware analysts can significantly reduce the burden of answering these two questions by exploiting the fact that most new malware is very similar to previously known malware. If we already had a database of known malware, we could essentially do a nearest neighbor search using our file similarity metric to query for known files that are similar to our new unknown file. If, for example, the most similar files in the database are malicious, we may assume that the new file is also malicious. And if many of the similar files in the database are a particular type of malware, we may also assume that the new file is that same type of malware.
 
\subsection{Fuzzy Hashing}
In the computer security community, file similarity is commonly measured using fuzzy hashing functions. A typical fuzzy hashing algorithm has two parts: (1) a hashing algorithm to generate a hash digest (e.g., the hash value representing a file), and (2) a comparison algorithm to compute a similarity metric from a pair of input hash digests (Bass et al. 2012).

Fuzzy hashing algorithms differ from cryptographic hash functions (e.g. MD5, SHA256, etc.) in that the output of a fuzzy hash function is insensitive to small changes in the input. For example, changing a single bit in the input to SHA256 will change the output drastically. However, with typical fuzzy hash functions, changing a single bit will not affect the output. Ideally, fuzzy hash functions should be as robust as possible to adversarial and random perturbations.

A well-known fuzzy hash function is ssdeep (Kornblum 2006). Given a file, ssdeep first breaks it up into several pieces, and then uses a rolling hash function to compute the digest for each piece. These individual digests are concatenated to produce a final \textit{fixed-size} similarity digest for the whole file. Given two hash digests, ssdeep will return a score of between 0 and 100 that expresses a level of confidence that two files are similar.

Despite ssdeep's popularity, it has several known weaknesses. For example, ssdeep is sensitive to random bit flips made across an entire file. In fact, if more than 70 bytes of a file are modified, ssdeep will always return a similarity score of zero when given the original and modified files as inputs (Breitinger 2011).

sdhash (Roussev 2010) is another widely used fuzzy hash function with increased robustness to random modifications of the input compared to ssdeep: sdhash can measure that a file is similar to a modified version of that file even when up to 1.1\% of the file has been modified at random. Unlike ssdeep, sdhash generates varying length digests (2-3\% of the length of the input). sdhash uses normalized Shannon entropy to extract 64-byte sequence features. The extracted features are hashed and then put into a Bloom filter (Bloom 1970). For the comparison of two files, Bloom filters are compared using a Hamming distance measure. The result is the estimated fraction of Bloom filter features that the two filters have in common that are not due to chance.

The traditional fuzzy hash function that is most similar to our work is sahash (Breitinger 2014). The simplistic design of sahash allows for its similarity score to be a lower bound on the Levenshtein distance. However, sahash is constrained to only fixed-sized inputs and cannot measure the similarity digests of files with different sizes. sahash works by concatenating the following sub-hashes: 1) the hash of the frequency of byte-arrays of files, 2) the hash of the frequency of 4-bit circularly shifted byte-arrays of files 3) a so-called \texttt{uneva} function. \texttt{uneva} not only counts the frequency of occurrences in the byte-arrays, but it also counts the frequency of the repeated occurrences, as a measure of how sporadic the occurrences of a specific byte are. We refer the reader to (Breitinger et. al 2014) for a more detailed description. Since we train our model using sahash features, we describe sahash features more in the feature preparation section.

We don't intend to provide a panacea to the fuzzy hash problem. The main novelty of our work is to propose a \textit{minimax training} framework, where fuzzy hash functions can be \textit{learned}. We show that learned fuzzy hash functions are more robust than popular traditional fuzzy hash functions to modifications of the inputs. We hypothesize that this robustness is due to traditional fuzzy hash functions being data agnostic, i.e., they don't learn from data, whereas our fuzzy hashing function can exploit learned statistical relationships in the underlying dataset. We also show that the tolerance of the fuzzy hashing function to file modifications can be implicitly learned from the corpus of data on which the fuzzy hashing function is trained. 

Note that we focus on fixed-size digests to have a \textit{differentiable end-to-end} mechanism for learning. This choice of fixed-sized digest helps us to do a fair comparison between ssdeep, sahash, and our method.

For our dataset, we start with a collection of binary files, and then modify the bytes of these files at random using a fuzzer, generating a new collection of perturbed files. Note that, while we use the term \textit{perturbed} in our paper to refer to these fuzzed examples in the training data, the term `perturbed' is relative, and the original files can be also considered to be the perturbed versions of the fuzzed files. We'll describe the dataset later on in the paper in our Experiments section.

Note that while we focus on applications of fuzzy hashing to file similarity for the purposes of malware analysis, fuzzy hashing can be applied to many kinds of data, including images and text.

\subsection{Fuzzy Hash Functions as Kernel Networks}
We view fuzzy hash functions as kernel networks. Inspired by (Goodfellow 2014) (Bo 2014) (Azarafrooz 2017), we incorporate kernel networks into a novel minimax framework. Our proposed minimax framework shares some similarities with cycle-consistent minimax frameworks (Zhu 2017) (Yi 2017). However, our approach is different in two ways: 1) The players of minimax games are kernel networks rather than neural architectures, and 2) Consistency is expressed in terms of the roles of the kernel networks w.r.t to the perturbed data. We refer to this as `perturbation consistency' because networks alternate their input data in each stage of learning, and whichever network uses the perturbed data as input will play the role of an adversary. We found such consistency to be crucial for satisfying the end-to-end symmetrical properties of a learned fuzzy hash function (we explain this further in the Learning Fuzzy Functions section). 

\section{Feature Preparation}
 Our proposed methodology works at the byte level and, therefore, data preparation requires no special parser. Given a file with byte-array \texttt{f}, we generate a perturbed file with byte-array $\texttt{f'}$ using random $\rho$-bit \textit{substitutions} s.t.  $0<\texttt{f}\oplus\texttt{f'}<\rho$, without changing the physical size or the runtime behavior of files, where $\oplus$ is \texttt{XOR} operator. This is not to say that a file is randomly selected from Hamming sphere of radius $\rho$, since some points in the Hamming sphere do \textit{not} remain truthful to the functionality of \texttt{f}. In other words, we avoid flipping bits that are essential to the functionality of the underlying executable files. The choice of using a bit substitution operation is to guarantee that the functionality of the perturbed version remains intact. Enabling perturbations with other edit operators, \textit{deletion} and \textit{insertion}, requires special care to avoid changing an executable file in such a way that it is no longer able to be executed. As explained later, we prepare test data in which files are fuzzed using deletion and insertion operations.
 
 Similar to sahash, for each byte-array \texttt{f}, we prepare features \texttt{x} by concatenating the histogram of byte counts with the histogram of byte counts after a circular shift:
 \begin{eqnarray}\label{hist}
\begin{array}{l}
\texttt{x} = \texttt{ hist(f) | hist(f >> 4)} 
\end{array}
\end{eqnarray}
, where \texttt{f >> n} denotes \texttt{n}-bit circular shift to the right, \texttt{hist(f)} is the histogram of byte counts and $|$ is the concatenation operator. Since each byte can be represented as a value between \texttt{0} and \texttt{255}, \texttt{x} is a \texttt{512} dimensional vector. For example, the zero-based indexes of \texttt{255} and \texttt{511} represent the counted numbers of byte occurrences of \texttt{0xff} in the original bytes and circularly shifted bytes, respectively. A more aggressive feature selection is also possible, where histograms of byte counts for other shift values $\texttt{n}: \texttt{n} \mod 8 \neq 0$ are incorporated into the features. Similar feature selection is done for the perturbed byte-array \texttt{f'} to get \texttt{x'}. The pairs (\texttt{x,x'}) $\in$ \texttt{X,X'} are then used to learn fuzzy hash functions.
  
\section{Learning Fuzzy Functions}

sahash (Breitinger et. al 2014) digest gets calculated as:

\begin{eqnarray}\label{sahash}
\begin{array}{l}
\texttt{x}\mod \texttt{m} ~~~ \text{where} ~~~ \texttt{m}=2^{\max(8,\ceil{\frac{\log_2 \texttt{l}}{2}})}
\end{array}
\end{eqnarray}
  
, where $\texttt{l}$ is the length of $\texttt{f}$.  The similarity measure for the sahash digests is then related to the maximum number of byte frequencies that are different among both the circulated and the original byte-array. This simply provides a lower bound for the edit distance between the files. Note that edit distance between two files is defined as the minimum number of substitutions, insertions, and deletions that are required to transform one file to another. The value of modulus \texttt{m} is selected to be file-size dependent for a suitable balance between compressibility and robustness. This, however limits its applicability to different file sizes. Instead of file-size dependent modulo operations, we learn a high-order non-linear function from the training data pairs \texttt{X,X'}. By doing so, similarity measure between digests finds intuitive mathematical interpretation in terms of maximum mean discrepancy (MMD) measure (Gretton 2012). 

To incorporate learning mechanisms in fuzzy hash functions, we need to extend the fuzzy hash function properties to include the following:
\begin{itemize}
\item \textit{Differentiability}: The desire to learn from the structure of the data $\texttt{X,X'}=(\texttt{x}_1,\texttt{x}'_1),(\texttt{x}_2,\texttt{x}'_2),...,(\texttt{x}_n,\texttt{x}'_n)$  asks for a differentiable network. For example, we cannot use the modulo operators used by sahash.
\item \textit{End-to-end symmetrical property}: We use a minimax framework with two different functions $(G, D)$ to compute the similarity measure $\delta$. This, however, requires an end-to-end symmetrical behavior, formalized below:
\begin{eqnarray}\label{symmetry}
\begin{array}{l}
\delta(G(\texttt{x}),D(\texttt{x'}))=\delta(G(\texttt{x'}),D(\texttt{x}))
\end{array}
\end{eqnarray}
Note this is more specific than the symmetrical properties of similarity/distance measures. The MMD similarity measure used in our paper is symmetric by default. 

\end{itemize}
 
To achieve the introduced properties above and those introduced in the introduction section, we model the fuzzy hash function selection using kernel minimax frameworks. This leads to differentiable non-linear embeddings. We view these embeddings as fuzzy hash digests which turn out to be robust against perturbations, thanks to the qualities of MMD distance measure and the minimax training. In order to achieve the end-to-end symmetrical property, a perturbation-consistent version of the minimax framework is presented.

\subsection{Kernel embedding in a Minimax Framework}
Let (\texttt{h},\texttt{h'}) denotes the digest of pairs (\texttt{x},\texttt{x'}) with similarity measures of $\delta(\texttt{h},\texttt{h'})$. We view the digests in our framework as the kernel embeddings of (\texttt{x},\texttt{x'}). This view establishes the connection between the required similarity measures of the fuzzy hash functions and maximum mean discrepancy (MMD) measures (albeit a stochastic version). The key idea behind MMD measure is the fact that one can measure the difference between distributions $p(\texttt{X})$ and $p(\texttt{X'})$ by linear witness in Hilbert space such as:

 \begin{eqnarray}\label{mmd-minimax}
\begin{array}{l}
~ \underset{f \in \mathcal{H}}{\mathrm{sup}} ~  \commentedbox{\mathbb{E}[\psi(\texttt{X})] -\mathbb{E}[\psi(\texttt{X'})]}{$\delta(\texttt{h},\texttt{h'})$ }
\end{array}
\end{eqnarray}
 
, without using density estimators over the original domains (\texttt{X},\texttt{X'}), where $\psi$ is a function living in a \textit{reproducing kernel Hilbert space} (RKHS) $H$.
Conveniently, in the process, we get the kernel embedding as the digest of the fuzzy hash functions. This provides digests and similarity measures in one shot. 
We further achieve \textit{learnable parameters} and robustness by incorporating the kernel networks into minimax frameworks followed by a simple mathematical trick introduced in (Bo 2014) (Azarafrooz 2017): take the derivative of Eq. \ref{mmd-minimax} to get gradients and then reconstruct a new network with fresh learnable parameters.

Before we continue with demonstrating the derivation of the new network with learnable parameters, we have to touch on a few definitions and one main theorem.

\textbf{Definition 1} (kernel mean embedding). Due to the \textit{reproducing property} of $H$, the expectation of any function $\psi$ in RKHS $H$ with respect to random variable \texttt{X} can be computed as an inner product with its so called {\it kernel mean embedding $\mathbb{E}[k(\texttt{X},.)]$}:
\begin{eqnarray}\label{reproducing-property}
\begin{array}{l}
\mathbb{E}[\psi(\texttt{X})]=\langle \psi, \mathbb{E}[k(\texttt{X},.)] \rangle
\end{array}
\end{eqnarray}
where $k$ is the kernel associated with RKHS $H$.

 \textbf{Definition 2} A kernel $k(\texttt{x},\texttt{x'}):\texttt{X}\times\texttt{X} \rightarrow \mathbb{R}$ is positive definite (PD), when for all $n>1$ and $x_1,x_2,..,x_n \in \texttt{X}$ and $c_1,...,c_n \in  \mathbb{R}$, we have $\sum_{i,j}c_ic_jk(\texttt{x}_i,\texttt{x}_j) \geq 0$.

\textbf{Theorem 1} \textit{Duality between Kernels and Random Processes} (Devinatz 1953):  If $k(\texttt{x,x'})$ is a positive definite kernel, then there exits a set $\Omega$, a measure $\mathbb{P}$ on $\Omega$, and random function $\phi_{\texttt{W}}(\texttt{x}):\texttt{X}\rightarrow \mathbb{R}$ from $L_2(\Omega,\mathbb{P})$, such that $k(\texttt{x,x'})=\int_{\Omega} \phi_{\texttt{W}}(x)\phi_{\texttt{W}}(x')d\mathbb{P}(\texttt{W})$.

\subsubsection{Reconstructing a minimax network}

Using Eq. \ref{mmd-minimax} and Eq. \ref{reproducing-property}, we note that gradient of the similarity measure is:

\begin{eqnarray}\label{derivative}
\begin{array}{l}
\frac{\partial \delta}{\partial \psi}= {\mathbb{E}} [k(\texttt{X},.)]-{\mathbb{E}} [k(\texttt{X'},.)]
\end{array}
\end{eqnarray}
 
 Applying Theorem 1 to Eq. \ref{derivative}, followed by Monte Carlo approximation analogous to random Laplace feature maps (Rahimi 2007), we arrive at the following gradient terms:

\begin{eqnarray}\label{fmap}
\begin{array}{l}
\tilde{\phi}(\texttt{x})-\tilde{\phi}(\texttt{x'}),  ~ ~ \tilde{\phi}(\texttt{x}) \equiv  \frac{\exp(-\texttt{x}^T\texttt{w}_1)...exp(-\texttt{x}^T\texttt{w}_\texttt{s})}{\texttt{s}}
\end{array}
\end{eqnarray}

 where $\texttt{w}_j$ are drawn from a proper distribution. In our paper, it is drawn from Gaussian $\mathbb{P}(\texttt{w})=\frac{\exp(-\|\texttt{w}\|_2^2/2)}{(2\pi)^{d/2}}$. We also investigated Laplace features maps (more on this can be found in the Appendix).

 Now we are in the position to construct a new parametrized learnable framework by the linear combination of the first-order gradient terms in Eq. \ref{fmap}. To implement this, we treat the gradient terms as the activation of the first layer in our network (note the non-linearity of the kernel embedding) followed by one more layer with softmax activation. We use batch normalization immediately after the first layer followed by dropout during training. This completes the design description for both involved networks in the minimax training framework, the generator $G$ and adversary $D$.
 
  In the well-known minimax process used by GAN (Goodfellow 2014), a generator tries to sample from a desired probability space, given random noise as input. An adversary then evaluates the generated sample by comparing it against the real data, according to a suitable distance measure, such as Jenson-Shannon, Wasserstein, MMD, etc.  However, in our case, the inputs for both generator and adversary belong to the space of training data \texttt{X,X'}. Parameters $\theta$ and $\phi$ get adjusted by the gradient flowing through both networks, in every stage of the game.
\begin{figure}[h]
\includegraphics[width=6cm]{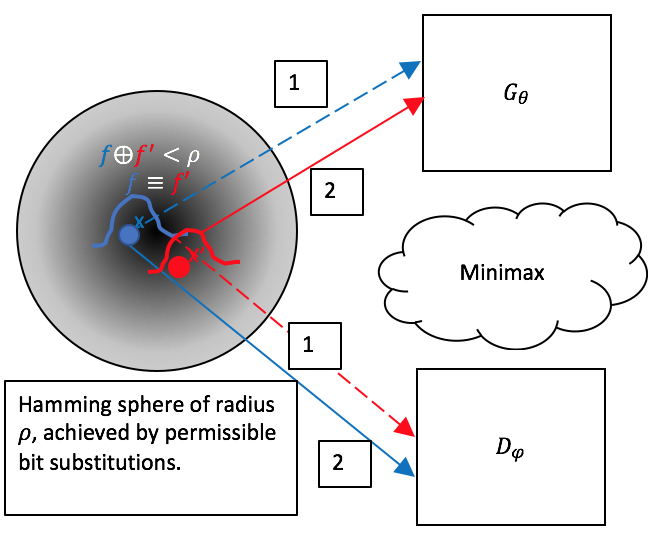}
\caption{Scheme of the proposed perturbation-consistent minimax training: Every stage of learning includes 2 rounds. Round  \commentedbox{1} ~~: $G_{\theta}$ uses \texttt{x} as input data, $D_{\phi}$ uses \texttt{x'}. $G_{\theta}$ plays the role of the generator and $D_{\phi}$ plays the role of adversary. Round  \commentedbox{2}  ~~ :  $D_{\phi}$ uses \texttt{x} as input data, $G_{\theta}$ uses \texttt{x'}. $D_{\phi}$ plays the role of the generator and $G_{\theta}$ plays the role of adversary. Both $G_{\theta}$ and $D_{\phi}$ backpropagate gradients to adjust their parameters, $\theta$ and $\phi$. Learning continues similarly with the next batch of data until convergence is achieved. }
\end{figure}

\begin{eqnarray}\label{Discrepency_optimization}
\begin{array}{l}
\underset{\theta  \in \Theta}{\mathrm{\min}} ~ \delta(G_{\theta} (\texttt{x}),D_{\phi} (\texttt{x'}))
\end{array}
\end{eqnarray}
   
  If the minimax characteristic of the framework is not clear from Eq. \ref{Discrepency_optimization} please refer to the definition of MMD distance in Eq. \ref{mmd-minimax} (find sup/max in Eq. \ref{mmd-minimax} and min in Eq. \ref{Discrepency_optimization}).
 The embedded features of either $G_{\theta}$ or $D_{\phi}$ can be considered the hash digest of resulting fuzzy hash function, with the similarity measure being the utility of the minimax game in \textit{equilibrium}.  
   
  To induce end-to-end symmetry, we train in a \textit{perturbation-consistent} way, where the role of players in the minimax training (i.e discriminator versus generator) \textit{alternates} along with their input data. In other words, during each state of training, we alternate input data to both networks, and whoever has access to the perturbed data \texttt{X'} in that round, plays the role of an adversary. This is visually illustrated in Fig. 1. We refer to our proposed method as \textit{adversarial kernel hash (akash)}.

\section{Experiments}
 \textbf{Data Preparation} To show that learning from data can result in an improved fuzzy hash function relative to traditional fuzzy hashing, we base our study on a set of Portable Executable (PE) files. The PE file format describes a data structure that encapsulates the information necessary for the Windows OS loader to manage the wrapped executable code. We curated a training set of \texttt{80} thousand malicious PEs, fuzzed using the aforementioned $\rho$-bit substitutions mechanism with $\rho$ selected uniformly random from range \texttt{1} to \texttt{500}.
 We generate two types of test data. One set of test data is generated using the explained flip substitution, while the other set is generated using LIEF (Quarkslab 2017). LIEF (Library to Instrument Executable Formats) is an open source library capable of parsing and manipulating PE files using advanced edit operations. 
 
 We avoided using LIEF for preparing the training data set for two reasons: 1) In order to show that even training over the Hamming sphere of permissible bit-substitutions generalizes to deletion and insertion edit operations. 2) Advanced edit operations using tools like LIEF can still sometimes generate a fuzzed PE file that is corrupted in such a way as to be unexecutable. However, our bit substitution mechanism is guaranteed to keep the functionality of the PE intact.  
 
 
 \textbf{Hyperparameter tuning}
 The best results were achieved by setting batch size to \texttt{1000}, dropout to \texttt{0.75} and learning rates to \texttt{5e-4}, for both generator and adversary networks. We found our experiments to be sensitive to the batch size. Also, we set the embedding size to \texttt{512}, without performing any tuning, in order to keep the digest size consistent with sahash. For optimization, we used (adaptive moment estimation) Adam stochastic optimization. We trained the model for \texttt{5000} epochs.

\textbf{Thresholds}  sahash uses the absolute value of the difference between the \texttt{uneva} of the files in order
to reduce false positives. While it is possible to incorporate the \texttt{uneva} arrays into input data \texttt{x}, we chose not to do so, in order to have a fair comparison between akash and sshash.
Similar to sahash, we call two files similar if they fall below a certain threshold for $\delta$ (Eq. \ref{mmd-minimax}) \textit{and} above a threshold for \texttt{uneva}. While sahash selected the threshold to optimal value of \texttt{97}, we achieved zero false positives by smaller value of threshold \texttt{80}. 
 
 \textbf{Visual Comparisons of Robustness}  
 To compare the robustness of akash with other fuzzy hash techniques, we plotted the effect of various types of perturbations on the distance measures in Fig. 2 and Fig. 3. In all these plots, the \texttt{y} axis in ssdeep and sdhash are (\texttt{100}-confidence score of finding files similar) and (\texttt{100}-estimate of the fraction of common features between files) respectively, as a proxy for their distance measures.  
 The \texttt{y} axis in sahash provides a lower bound estimate on the edit distance. The \texttt{y} axis in akash is representative of distance measure $\delta$, described in Eq. \ref{mmd-minimax} as the utility of the minimax game at equilibrium. The \texttt{x} axis in Fig. 2 represents the number of bit-substitutions performed in randomly selected files. However, the modified file has the same physical size and functionality as the unmodified file. Disregarding the plot scales, the large spikes in the figures correspond to the events where fuzzy hash functions are being bypassed, i.e., where the fuzzy hash functions fail to consider two similar files to be similar. Lower numbers of spikes implies more robustness. It can be observed that akash is consistently more robust than the other fuzzy hash functions.  Despite training on only $\rho$-bit substitutions within $0<\rho<500$, Fig. 2 demonstrates that akash's performance generalizes to far more than 500 modifications.
  \begin{figure}[h]
\includegraphics[scale=0.5]{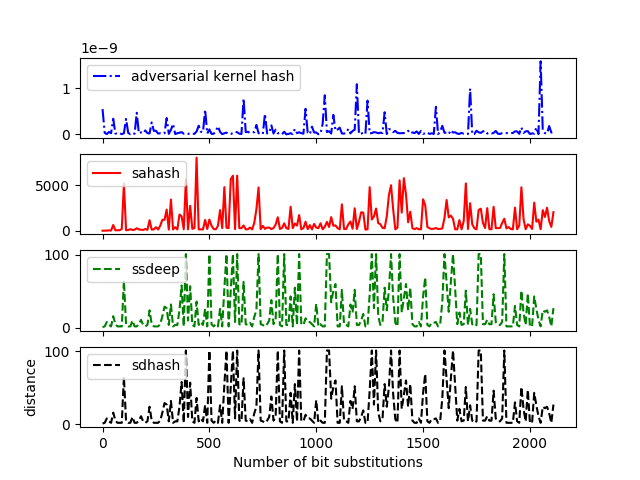}
\caption{The effect of increase in the number of (permissible) bits substitutions on the distance measures. The smoother the trends, the more robust the fuzzy hasher.}
\end{figure}

\begin{figure}[h]
\includegraphics[scale=0.5]{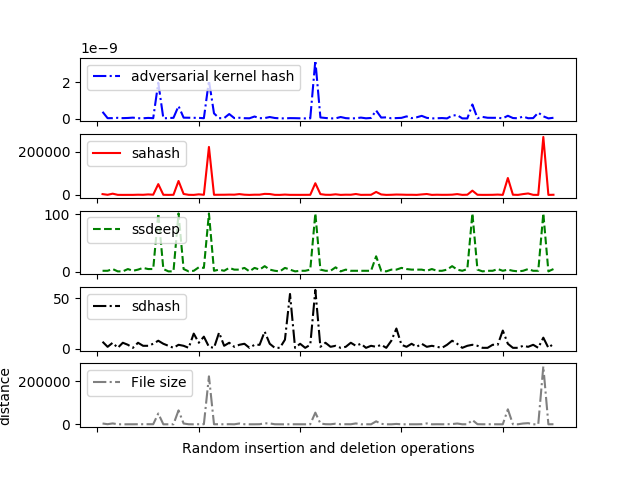}
\caption{The effect of random insertion/deletion operations on the distance measures. The smoother the trends, the more robust the fuzzy hasher. Notably, training over bit substitutions results in a fuzzy hasher that generalizes well to insertions and deletions.}
\end{figure}

More interestingly, Fig. 3 shows that akash generalizes well to even more advanced edit operations such as insertions and deletions.
The \texttt{x} axis in Fig. 3 represent the events with random action selected from the insertion and deletion actions available to LIEF.  The insertion operations include the following LIEF actions: `append overlay', `append import, `add section' and `append section'. The deletion operations include the following LIEF actions: `remove the signature' and `remove debug'. 

 We also plotted the byte size of the modifications associated with the performed edit actions.
Fig. 3 shows that akash is robust even to the cases where files are generated via deletion and insertion operators rather than bit substitutions.

\textbf{Performance} 
We note that sdhash (unlike akash, ssdeep, and sahash) generates varying length digests (2-3\% of the length of the input). Therefore, a fair comparison between sdhash and the rest of the mentioned fuzzy hashers is not possible, but we report the sdhash results to be comprehensive. Table 1 and Table 2 shows the result of comparison over a randomly selected set of 10 thousand benign and malicious hashes. We use the recommended threshold of \texttt{21} for sdhash, \texttt{50} for ssdeep and the exact thresholds of sahash reported in (Breitinger et. al 2014).  \textit{Table 1 shows that akash outperforms the traditional fuzzy hashing functions, when perturbations are induced using bit-substitution operations}. Table 2 shows that, despite being trained on the space of permissible bit-substitutions, akash generalizes well even to the cases where perturbations are induced using insertions and deletions. 

 \begin{table}
\caption{True Negative Ratio: \textbf{bit-substitions}}
\begin{center}
\begin{tabular}{ |c|c|c|c|c| } 
 \hline
 Fuzzy Hash & \textit{akash} & ssdeep & sahash & sdhash \\
 \hline
 \hline
TN \% & \textit{97.9} & 86.3 & 31 & 93.4
\end{tabular}
\end{center}
\end{table}
 
   \begin{table}
\caption{True Negative Ratio: \textbf{Insertion and deletion}}
\begin{center}
\begin{tabular}{ |c|c|c|c|c| } 
 \hline
 Fuzzy Hash & \textit{akash} & ssdeep & sahash & sdhash \\
 \hline
 \hline
Overall \% & \textit{85} & 96.3 & 28 & 90.1 \\
Insertion \% & \textit{88} & 96 & 42 & 93.4 \\
Deletion \% & \textit{75} & 99 & 36 & 77
\end{tabular}
\end{center}
\end{table}

 \begin{table}
\caption{True Negative Ratio for Compression Operations}
\begin{center}
\begin{tabular}{ |c|c|c|c|c| } 
 \hline
 Fuzzy Hash & \textit{akash} & ssdeep & sahash & sdhash \\
 \hline
 \hline
UPX \% & \textit{0} & 20 & 20 & 14.2 
\end{tabular}
\end{center}
\end{table}

 \begin{table}
\caption{True Positive Ratio}
\begin{center}
\begin{tabular}{ |c|c|c|c|c| } 
 \hline
 Fuzzy Hash & \textit{akash} & ssdeep & sahash & sdhash \\
 \hline
 \hline
TP \% & \textit{100} & 98.3 & 98.3 & 100
\end{tabular}
\end{center}
\end{table}

We also experimented with more advanced edit operation such as compression. Specifically, we used the UPX pack/unpack compression action in LIEF when conducting this new experiment, and found zero detections in this case, as is reported in Table 3. This result is expected, since compression changes the structure of the original files in much more substantial way than substitution, insertion, and deletion operations.

 \textbf{False Positives} In order to measure the false positive rate, we selected a random set of \texttt{n=10000} malicious hashes. We then selected a random selection of \texttt{10000} pairs from the possible set of combinations $\texttt{n} \times \texttt{n-1}/2$. The results are reported in Table 4. We performed similar experiments for a set of benign hashes. The numbers are consistent with Table 4 as well.  We found the true positive ratio for both sahash (using the same exact settings in the original paper) and ssdeep to be the same: 98.3\%, which is lower than the perfect 100\% performance of akash and sdhash.  Results in Table 4 shows that akash performs as good as sdhash, in terms of ratio of false positives, despite having fixed-size digest.
 
\section{Conclusion and Future Work}
 We introduced a novel minimax training framework capable of learning to perform fuzzy hashing for the purposes of file similarity. The kernel embeddings of the minimax players can be used as hash digests, and the utility of the minimax game in equilibrium provides the similarity score. As shown in our experiments, the trained fuzzy hash function outperforms widely used traditional fuzzy hash functions when perturbations are induced using bit substitutions. Also, despite being trained on the space of permissible bit substitutions, the trained fuzzy hash function generalizes well even to the cases where perturbations are induced using insertions and deletions. 

We ran our experiments with PE files, but we suspect our approach can be fruitfully applied to other kinds of data, such as images or audio. For example, one could modify a collection of images using cropping, color changes, and geometric transformations, and then use our technique to measure the similarity of the modified and original images. 

For files in particular, an interesting future direction is to investigate to what extent a trained model is generalizable to perturbations involving more advanced edit operations, such as compression, instead of simple bit substitutions.



\subsection{References} 

\smallskip \noindent AV-TEST Security Report 2017/2018. 2018. Magdeburg, Germany. AV-TEST GmbH.

\smallskip \noindent Azarafrooz, M. 2017. Doubly Stochastic Adversarial Autoencoder. \textit{Second Workshop on Bayesian Deep Learning, NIPS 2017}: Long Beach, Calif.

\smallskip \noindent Baier, H. and Breitinger, F. 2011. Security Aspects of Piecewise Hashing in Computer Forensics. In \textit{Proceedings of the Sixth International Conference on IT Security Incident Management and IT Forensics}, 21-36. IEEE.

\smallskip \noindent Bass, L., Brown, N., Cahill, G., Casey, W., Chaki, S., Cohen, C., Niz, D. de, French, D., Gurfinkel, A., Kazman, R., Morris, E., Myers, B., Nichols, W., Nord, R.L., Ozkaya, I., Sangwan, R.S., Simanta, S., Strichman, O., Valetto, P., 2012. Results of SEI Line-Funded Exploratory New Starts Projects, Technical Report. CMU/SEI-2012-TR-004, ESC-TR-2012-004, Software Engineering Institute, Carnegie Mellon Univ., Pittsburgh, Penn.

\smallskip \noindent Bloom, B.H.. 1970. Space/Time Trade-Offs in Hash Coding with Allowable Errors. \textit{Communications of the ACM} 13(7): 422-426.

\smallskip \noindent Breitinger, F., Ziroff G., Lange, S., Baier, H. 2014. Similarity Hashing Based on Levenshtein Distances. In \textit{Advances in Digital Forensics X. DigitalForensics 2014. IFIP Advances in Information and Communication Technology}, vol. 433. Berlin, Heidelberg: Springer.

\smallskip \noindent  Dai, B., Xie, B., He, N., Liang, Y., Raj, A., Balcan, M.F.F. and Song, L.. 2014. Scalable Kernel Methods via Doubly Stochastic Gradients. In \textit{Advances in Neural Information Processing Systems} 27: 3041-3049.

\smallskip \noindent Devinatz, A. 1953. Integral Representation of Positive Definite Functions. \textit{Transactions of the American Mathematical Society} 74(1) 56-77.

\smallskip \noindent Goodfellow, I., Pouget-Abadie, J., Mirza, M., Xu, B., Warde-Farley, D., Ozair, S., Courville, A. and Bengio, Y. 2014. Generative Adversarial Nets. In \textit{Advances in Neural Information Processing Systems}: 2672-2680.

\smallskip \noindent Gretton, A., Borgwardt, K.M., Rasch, M.J., Scholkopf, B. and Smola, A.. 2012. A Kernel Two-Sample Test. \textit{Journal of Machine Learning Research}, 13 (March 2012): 723-773.

\smallskip \noindent Kornblum, J.. 2006. Identifying Almost Identical Files Using Context Triggered Piecewise Hashing. \textit{Digital Investigation} 3: 91-97. 

\smallskip \noindent Quarkslab. 2017-2018. LIEF: Library for Instrumenting Executable Files. https://lief.quarkslab.com. Last accessed: Sept. 5, 2018.

\smallskip \noindent Rahimi, A. and Recht, B.. 2008. Random Features for Large-scale Kernel Machines. In \textit{Advances in Neural Information Processing Systems}: 1177-1184.

\smallskip \noindent Roussev, V. Data Fingerprinting with Similarity Digests. In \textit{Advances in Digital Forensics VI}, K. Chow and S. Shenoi, Eds., \textit{Springer}, Heidelberg, Germany, 207--226, 2010.

\smallskip \noindent Walenstein, A., and Lakhotia, A. 2007. The software similarity problem in malware analysis. In \textit{Dagstuhl Seminar Proceedings}. Dagstuhl, Germany: Schloss Dagstuhl-Leibniz-Zentrum for Informatik.

\smallskip \noindent Yi, Z., Hao (Richard) Zhang, Tan, P. and Gong, M.. 2017. Unsupervised Dual Learning for Image-to-Image Translation. In \textit{International Conference on Computer Vision (ICCV)}.

\smallskip \noindent Zhu, J.Y., Park, T., Isola, P. and Efros, A.A.. 2017. Unpaired Image-to-Image Translation Using Cycle-Consistent Adversarial Networks. In \textit{International Conference on Computer Vision (ICCV)}.

\subsection{Appendix}
 We also experimented with Laplace feature maps with samples drawn Exponential and Levi. But we found the Fourier features maps to perform much better than Laplace feature maps. The Laplace feature maps led to fuzzy hashers that have an impractical FP ratio. This is an interesting observation since semigroup kernels associated with Laplace feature maps provide a much more intuitive interpretation than Fourier Features maps. This is because one can consider histograms \texttt{X} as groups closed over the \texttt{+} operation. Since byte counts can only get positive values.  We did however, find that the convergence for Laplace Feature maps with samples drawn from exponential to be much smoother. 

\textbf{Definition 3} (Abelian semigroup). A semigroup (\texttt{X}, $\circ$) is a nonempty set \texttt{X} equipped with an associative composition $\circ$, i.e for any $\texttt{x,x',x''} \in
  \texttt{X} :  \texttt{x} \circ ( \texttt{x'} \circ  \texttt{x''}) =  (\texttt{x}\circ  \texttt{x'})\circ \texttt{x''}$ and a identity element \texttt{e}, i.e, for any  $\texttt{X} \in  \texttt{X}: \texttt{x}\circ  \texttt{e} = \texttt{x}.$ For an abelian semigropu, the composition is commutative, i.e. for any  $\texttt{x}, \texttt{x'} \in  \texttt{X}: \texttt{x} \circ  \texttt{x'} = \texttt{x'} \circ  \texttt{x} .$

\textbf{Definition 4} (Kernels on Abelian semigroups). A function $k: \texttt{X} \times \texttt{X} \rightarrow \mathbb{R}$ is a positive definite (PD) kernel function on an abelian semigroup $(\texttt{X},\circ)$ if $k(\texttt{x},\texttt{x'}) = \Phi(\texttt{x} \circ \texttt{x'})$ where $\Phi: \texttt{x} \rightarrow \mathbb{R}$ is a PD function, i.e for any $\texttt{x}_1,...,\texttt{x}_n \in \texttt{X}$, any real-valued scalars $\texttt{c}_1,...,\texttt{c}_n$, the following holds: $\sum_{i,j=1}^{n} \texttt{c}_i\texttt{c}_j\Phi(\texttt{x}_i,\texttt{x}_j)\geq 0$.

The following theorem establishes a one-to-one correspondence between semigroup kernels and probability densities on $\mathbb{R}^d$, via the Laplace transform.

\textbf{Theorem 1} (Duality between semigroup kernels and random processes):  For every bounded continuous kernel function $k(\texttt{x},\texttt{x'})$ on the Abelian Semigroup $(\mathbb{R}^d_+,+)$ there exits a non-negative measure \texttt{w} with associated \textit{random Laplace} construction $\phi_W$ s.t that $k(x,x')=\int_{\mathbb{R}^D_{+}} \phi_{\mathcal{W}}(x)\phi_{\mathcal{W}}(x')d\mathbb{P}(\mathcal{W})$.

\end{document}